\documentclass[conference]{IEEEtran}
\usepackage{stmaryrd}
\usepackage{amsfonts}
\usepackage{graphicx,times,amsmath} % Add all your packages here
\IEEEoverridecommandlockouts % to create the author's affliation portion
\begin{document}
% paper title: Must keep \ \\ \LARGE\bf in it to leave enough margin.
\title{\ \\ \LARGE\bf A General Framework for Development of the Cortex-like Visual Object Recognition System: Waves of Spikes, Predictive Coding and Universal Dictionary of Features\thanks{S. Tarasenko is with Department of Intelligence Science and Technology, Graduate School of Informatics, Kyoto University, under JST ERATO Asada Synergistic Intelligence Project (email: infra.core@gmail.com).} %\thanks{This work was supported by a CUHK Direct Grants under Project Code
%2050375}
}
\author{Sergey S. Tarasenko}
% avoiding spaces at the end of the author lines is not a problem with
% conference papers because we don't use \thanks or \IEEEmembership
% use only for invited papers
%\specialpapernotice{(Invited Paper)}
% make the title area
\maketitle
\begin{abstract}
This study is focused on the development of the cortex-like visual object recognition system. We propose a general framework, which consists of three hierarchical levels (modules). These modules functionally correspond to the V1, V4 and IT areas. Both bottom-up and top-down connections between the hierarchical levels V4 and IT are employed. The higher the degree of matching between the input and the preferred stimulus, the shorter the response time of the neuron. Therefore information about a single stimulus is distributed in time and is transmitted by the waves of spikes. The reciprocal connections and waves of spikes implement predictive coding: an initial hypothesis is generated on the basis of information delivered by the first wave of spikes and is tested with the information carried by the consecutive waves. The development is considered as extraction and accumulation of features in V4 and objects in IT. Once stored a feature can be disposed, if rarely activated. This cause update of feature repository. Consequently, objects in IT are also updated. This illustrates the growing process and dynamical change of topological structures of V4, IT and connections between these areas.
\end{abstract}
% no key words
\section{Introduction}
\label{intro}
\PARstart{V}{isual} object recognition systems have been broadly discussed in numerous studies from various fields. Originated from vision research, visual object recognition systems are now subjects for interdisciplinary studies. At present, this is the field, where artificial intelligence (AI) meets neuroscience.

One of the strongest recent trends is to create the systems, which exhibit biologically plausible performance. This purpose is achieved by tuning the parameters of the systems to the neurophysiological data. On the other hand, their architectures replicate the structure of the corresponding cortical systems.

According to neurophysiological findings, the ventral visual stream (VVS) is the brain system responsible for visual object recognition. The VVS includes the V1 area (primary visual cortex), the V2, V3 and V4 areas and the IT area (Inferior Temporal Cortex).

The VVS has a hierarchical bottom-up structure of visual areas. Along the VVS in the anterior direction starting from the primary visual cortex, the size of receptive fields (RFs) of neurons increase. Simultaneously, the selectivity of neurons decreases, i.e., if the V1 neurons are strictly tuned to a bar of a single orientation and of a particular size, the V4 neurons are tuned to various geometric forms and exhibit higher degree of translation and scale invariance. Consequently, a complexity and a size of visual features, recognized in the consecutive areas along the ventral stream, increase.

Therefore, most visual object recognition systems, born of a symbiosis of AI and neuroscience, have the functional architecture that resembles the hierarchical structure of the VVS. Although not all the visual areas of real VVS are employed by the models, usually V1 and IT areas are used. The area in-between is usually either V2 \cite{korner} or V4 \cite{riepog}.

Results of neurophysiological studies indicate that recognition task in the VVS is performed extremely fast. The processing time varies from 100-150ms for humans \cite{thim}, and it is even shorter for monkeys \cite{fabre}.

The extremely short processing time implies that no feedback (top-down) interaction occurs during visual processing. Therefore many researchers attempted to model the feature of high speed processing by implementing only feed-forward connections \cite{thorpe96,riepog,serre3}. 
Furthermore, rapid processing implies that a single neuron at each level fires only once, producing a \textit{single spike}\cite{thorpevan}(p. 716). 

On the other hand, the bottom-up (feed-forward) and top-down (feedback) processing circuits can be found throughout the cortex. Therefore, application of only feed-forward processing is a great simplification.
The reciprocal information processing in the brain has been efficiently used in the concept of \textit{predictive coding}, which has been introduced in \cite{kawato1,rao:bal}.

Most recently, a hierarchical network with reciprocal (bottom-up and top-down) connections has been presented \cite{korner}. This model employes propagation of waves of spikes and uses reciprocal connections to implement predictive coding. 

Although, numerous models of the VVS have been presented, the matter of development of visual areas in VVS was not broadly discussed. Several studies have addressed the matter of development of orientation selectivity in primary visual cortex V1 \cite{bedmik1,crairetal}, modular structures \cite{chermood} and inter-cortical development \cite{bedmik2}. The key mechanism of structure formation in developmental models is Hebbian learning \cite{chermood}, while growth is implemented by growing Self-Organizing Maps\cite{bedmik2}.
\begin{figure}
\centering
\includegraphics[width=8.5cm]{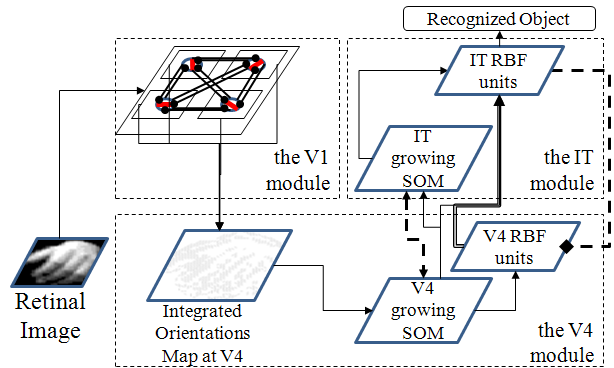}
\caption{Schematic representation of the system architecture. Please refer text for detailes.}
\label{sketch}
\end{figure}

We consider \textit{development} on the scale of the VVS as extraction of new visual features in the V4 area and new objects in IT. This is implemented by means of growing self-organizing networks \cite{fritzke}. Nevertheless, growing structure is not the only feature of the development. We also consider development to reflect the emergence of new connections between V4 and IT due to inclusion of the new instances (features in V4 and objects in IT). 

Therefore, development is regarded as a process of growing and establishing new connections of both intra- and inter-level types. In other words, development is the emergence of new neurons and new connections within and between hierarchical levels. The origin of development is based on extraction of new visual features across all hierarchical levels.

In this study, we introduce the framework to model developmental processes in the cortex-like visual object recognition system. 

\section{Functional Architecture of the System}
This framework consists of three processing levels (modules). These levels correspond to the V1, V4, and IT areas of the VVS. The V4 and IT levels can be also segregated into layers. Both V4 and IT levels have layers of growing SOM and Radial Basis Function (RBF) units. Meanwhile, the V4 level also has integrating layer. Each layer is a single neural network. The schematic representation of the entire system, including the connections between the levels (cortical areas) and layers (networks inside the cortical areas), is presented in Fig.~\ref{sketch}.

The V1 level contains four distinct groups of neurons (filter maps), which are mutually interconnected with inhibitory connection (round-head solid lines). After inhibition the information from all four maps is integrated by the layer of the V4 level. The output of this layer is \textit{Integrated Orientation Map (IOM)}. The IOM is then sent to the V4 growing SOM layers by means of excitatory connections. The excitatory connections are indicated by arrow-head solid lines throughout the schema. After IOM is translated into the Feature Map, it is sent V4 RBF units layer and to the IT growing SOM layer. The IT RBF units layer receives excitatory inputs from the IT growing SOM layer as well. The IT RBF units layer sends modulatory (amplification of existing activation level) signals (marked with diamond-head dashed lines) back to the V4 RBF units layer. The feedback modulated (amplified) signal from the V4 RBF units layer to IT RBF unit layer is marked with arrow-head double solid line. The V4 and IT growing SOM layers serve as repositories of visual features and objects, respectively. During the development the collection of visual features is changing. To update objects with valid features, interaction between the V4 and IT growing SOM layers emerges. This is indicated by double-end arrow-head dashed line.

The details of processing in each module, intra- and inter- layer and level intractions are discussed in subsequent sections. 

\begin{figure}
\centering
\includegraphics[width=8.5cm]{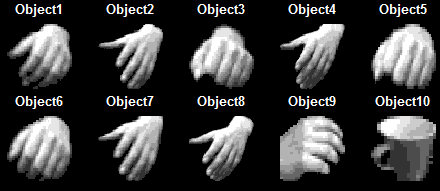}
\caption{Stimuli used in simulations.}
\label{stH} % Give a unique label
\end{figure}

For the purpose of simulation we use ten stimuli (objects) presented in Fig.~\ref{stH}. The stimuli are $100 \times 100$ pixel pictures of hands and cup. We use 11 black and white color gradations from 0 (black) to 1 (white) with step 0.1 on the normalized scale throughout the processing.

\subsection{Processing in the V1 area}
\label{v1proc}
In this study, we employ Gabor filters, as described by eq. (\ref{eqGab}), to model V1 simple cells: 
\begin{equation}
\begin{array}{l}
F(x,y) = \exp \left( { - \frac{{(x_o^2 + {\gamma ^2}y_o^2)}}{{2{\sigma ^2}}}} \right) \times \cos \left( {\frac{{2\pi }}{\lambda }{x_o}} \right) \\ 
{x_o} = x\cos \theta + y\sin \theta \\ {y_o} = - x\sin \theta + y\cos \theta \\ 
\end{array}
\label{eqGab}
\end{equation}
where $\theta$ is an orientation parameter, $\sigma$ is a scale parameter, and $\lambda$ is a wave length parameter. The equations for the Gabor filter in our study have been taken from \cite{serre3}.

We use a single value for the scale parameter: $\sigma = 2.8$, which corresponds to $7 \times 7$ pixels RFs. The relationship between scale and wavelength parameters are presented in \cite{serre1}(p. 69).

The input (retinal image) is processed with Gabor filters of four different orientations (0$^\circ$, 45$^\circ$, 90$^\circ$, and 135$^\circ$). Therefore in the V1 area there are four distinct groups of neurons with different orientations. The RFs of these cells `densely cover' the entire retinal image. Performance of each group is implemented as convolution of the Gabor filter kernel with the retinal image. The result of convolution is an orientation map. Each orientation map contains the response of the neurons selective for the same orientation, but situated at different locations.
To identify a particular orientation at each location, we employ all-to-all lateral intra-layer inhibitory connections to suppress the competing features in the V1 area \cite{korner}.
\begin{figure}
\centering
\includegraphics[width=8.5cm]{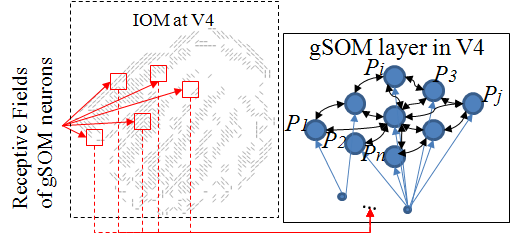}
\caption{Collection of prototypes $P_i$, $\{i=1,...,k\}$ with growing self-organizing maps, where $k$ is a total number of features. Not all the connection are shown.}
\label{gSOM} % Give a unique label
\end{figure}
\begin{figure}
\centering
\includegraphics[width=4cm]{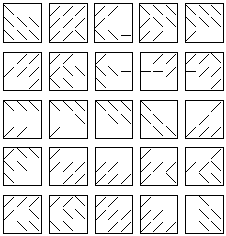}
\caption{The sample features (prototypes) collected by the gSOM from the input image (Object 1) in Fig.~\ref{stH}. The orientations are indicated with orientation bars. Each prototype is a $3 \times 3$ matrix of specified orientations. The empty space indicates absence of any orientation detected.}
\label{sampFeats} % Give a unique label
\end{figure}

\subsection{Processing in the V4 area}
\label{v4proc}
We use three computational types of V4 neurons. Our assumption of several types of neurons is supported by the recently proposed models \cite{riepog,buia}. 

Type I neurons integrate information from different orientation maps ($V4_{INT}$), resulting from mutual inhibition. The IOM (output of the $V4_{INT}$ neurons) is then processed to extract complex visual features as combinations of different orientations. To do so, the gSOM \cite{fritzke} is used. The neurons in gSOM are the V4 type II neurons ($V4_{gSOM}$).
The V4 neurons of type III are RBF-units \cite{pasucon}:
\begin{equation}
V4_{RBF} = exp(-\beta_{V4}|| X - P_i ||^2)
\end{equation}
where $\beta_{V4}$ is a tuning parameter, $P_i$ is the center of the RBF unit ($prototype$) and $X$ is the current input of the RBF-unit. A prototype is a visual feature (preferred stimulus) of a certain $V4_{RBF}$ neuron. 

In terms of Gaussian RBF, parameter $\beta_{V4}$ is the inverted doubled variance: $\beta_{V4} = \frac{1}{2Var^2}$. On the other hand, variance $Var^2$ of the Gaussian function is usually chosen as some fraction of the distance from the origin to the object in the feature space \cite{platt}. Such distance can be calculated as 
\begin{equation}
dist_{V4} = \sum\limits_{i = 1}^M \sum\limits_{j = 1}^M {\Theta_{ij}^2}, 
\end{equation}
where $\Theta_{ij}$ is a particular orienation at a given location of a visual feature, where $M$ and $N$, are vertical and horizontal dimensions of the IOM, respectively. $\Theta$ can take one of four integer values 1, 2, 3, or 4, which correspond to 0$^\circ$, 45$^\circ$, 90$^\circ$, and 135$^\circ$ orientation, respectively. The $Var^2$ is then taken to be one tenth of the $dist_{V4}$.

In our model, the RF size of $V4_{gSOM}$ is $3\times3$ matrix (RF-matrix) of the IOM elements. Concatenating the strings of the RF-matrix, we obtain 9-$dim$ vectors, which are used as inputs to the gSOM network.
\begin{figure}
\centering
\begin{tabular}{*{20}{c}}
\multicolumn{4}{c}{...}&{}&\multicolumn{4}{c}{...} \\
{...} & {28} & {32} & {...} & {} & {...} & {0.8} &{1} & {...} \\
{...} & {2} & {10} & {...} & {} & {...} &{1}& {0.4} & {...} \\
\multicolumn{4}{c}{...}&{}&\multicolumn{4}{c}{...} \\
%{...} & {...} & {...} & {...} & {} & {...} & {...} & {...}& {...} \\
\multicolumn{4}{c}{Sample part of}&{}&\multicolumn{4}{c}{Corresponding part of} \\
\multicolumn{4}{c}{Feature Map}&{}&\multicolumn{4}{c}{Response Map} \\
\end{tabular}
\caption{Sample part of Feature Map and corresponding part of Response Map.}
\label{matrs}
\end{figure}
\begin{figure}
\centering
\begin{tabular}{*{20}{c}}
\multicolumn{4}{c}{...}&{}&\multicolumn{4}{c}{...} \\
{...} & {0} & {1} & {...} & {} & {...} & {0} & {32} & {...} \\
{...} & 1 & {0} & {...} & {} & {...} & 2 & {0} & {...} \\
\multicolumn{4}{c}{...}&{}&\multicolumn{4}{c}{...} \\
\multicolumn{4}{c}{Sample First Wave}&{}&\multicolumn{4}{c}{Sample First Wave} \\
\multicolumn{4}{c}{Response Map}&{}&\multicolumn{4}{c}{Feature Map} \\
\end{tabular}
\caption{First Wave Feature and Response Maps. Zero value in Feature Map means absence of feature with unit activation at this location.}
\label{fwResp}
\end{figure}

The information about the features is stored in synaptic weights, therefore gSOM serves as feature repository. We do not use any updates for synaptic weight vectors, but only record the inputs that are sufficiently far from each other (in terms of Euclidean distance between the 9-dim synaptic vectors: if the distance between the input and $V4_{gSOM}$ neuron is greater than $0.1 \ dist_{V4}$, input is considered to be distant enough).

Each $V4_{gSOM}$ neuron plays a role of RBF-center (a prototype) for $V4_{RBF}$. Therefore, if $k$ is a total number of $V4_{gSOM}$ neurons, there are $k$ different classes of $V4_{RBF}$ neurons. The $V4_{RBF}$ neurons of a certain class cover densely the IOM. The RF size of $V4_{RBF}$ is identical to the RF size of the $V4_{gSOM}$ neurons. Thus, the total number of $V4_{RBF}$ neurons is then $kL$, where $L$ is the total number of neurons required to densely cover IOM.

The process of on-line visual feature extraction is illustrated in Fig.~\ref{gSOM}. Here sample RFs of gSOM neurons, densely covering the IOM, are presented. The $3\times3$ features are processed with gSOM, and a collection of distant enough visual features is extracted. The sample features extracted from the first stimulus are presented in Fig.~\ref{sampFeats}.

\textit{Processing the Integrated Orientation Map with $V4_{RBF}$ neurons: Feature and Response Maps.} After the features have been collected, the visual stimulus is represented as a \textit{Response Map}. To do so, we first assemble the \textit{Feature Map} of the visual stimulus by finding the feature that best matches at each spatial location on the IOM. Therefore, the \textit{Feature Map} is a $m \times n$ matrix, where $m < M$ and $n < N$, are vertical and horizontal dimensions, respectively. The Feature Map contains numbers of corresponding prototypes that best match the actual visual feature, presented at a particular spatial location. Then the responses of $V4_{RBF}$ neurons are calculated. This way, we obtain the \textit{Response Map}. 

The Response Map is also $m \times n$ matrix, containing responses of $V4_{RBF}$ neurons at each location of the IOM.

The sample parts of the Feature and Response Maps are presented in Fig.~\ref{matrs}.
\subsection{Waves of Spikes: Information Transfer from the V4 area to the IT area}
The neurons, whose preferred stimulus is closer to the actual visual input, fire faster \cite{van1}. Therefore, the neurons with unit activation (response is 1 on the normalized scale) fire first. Then the neurons with activation greater or equal to 1-$\epsilon$ ($\epsilon = 0.1$), excluding previously fired neurons, will fire and so on. Thus, the neurons with the same activation level form `waves of spikes'. Consequently, the original retinal image is unfolded in a time domain. Thus, information about the stimulus is carried by the sequence of waves of spikes.

To extract the \textit{the First Wave of Spikes}, we eliminate from the Feature Map the features that resulted in neural activation levels less than 1 on the normalized scale. The resultant map is called a \textit{First Wave Feature Map} (Fig.~\ref{fwResp}). This map consists only of the features that identically matched the actual visual stimulus. 

\begin{figure}
\centering
\includegraphics[width=7cm]{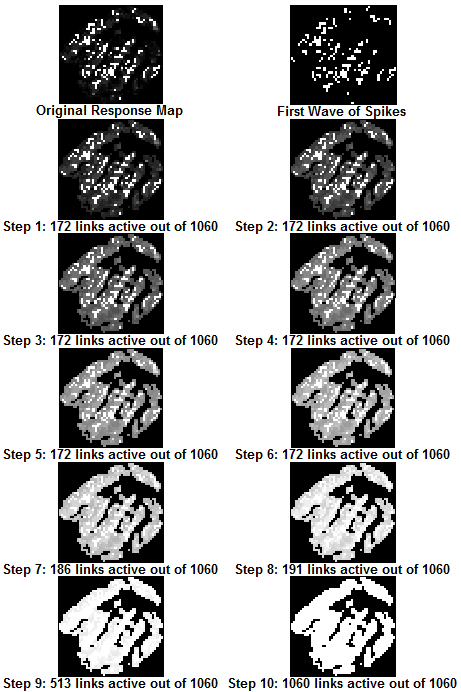}
\caption{The waves of spikes corresponding to the first stimulus. In \textit{Original Response Map} and the \textit{First Wave of Spikes} white color indicates unit (of magnitude 1) response, while black color indicates zero responses. In the Steps 1 to 10, white color indicates the neuronal signals that reached IT, i.e., waves of spikes that reached IT. In each step, the cumulative amount of spikes reached IT is indicated. For example, Step 6 illustrates the neurons, whose neuronal signals have already reach IT, i.e., six waves of spikes have arrived. Term `links' means bottom-up inter-level connection between V4 and IT neurons.}
\label{wh1} % Give a unique label
\end{figure}

The waves of spikes corresponding to the first stimulus are presented in Fig.~\ref{wh1}.

\subsection{Processing in the IT area}
Some of inferior temporal cortex neurons can be considered as RBF-units \cite{logpog}. The IT area is also addressed as a storage place of various objects and is suggested to have a function of object recognition and classification \cite{tanaka1}.

The objects in the IT area are obtained through every-day visual experience. Therefore, the structure of IT changes in time \cite{tanaka1}. To model this change, we applied gSOM to simulate the developmental activity of the IT area. The $IT_{gSOM}$ neurons are the IT neurons of type I. The activity of the gSOM network is similar to the V4 gSOM network.

The neurons of type II are $IT_{RBF}$ neurons. The $IT_{gSOM}$ neurons store the objects themselves, while $IT_{RBF}$ neurons perform matching between the object stored by the corresponding $IT_{gSOM}$ neuron (preferred stimulus) and the actual stimulus (input):
\begin{equation}
IT_{RBF} = exp(-\beta_{IT}|| X - Obj_i ||^2)
\end{equation}
where $\beta_{IT}$ is a tuning parameter, $Obj_i$ is the center of the RBF-unit and $X$ is the current input. The tuning parameter $\beta_{IT}$ estimated in the same way as it is done for $V4_{RBF}$ neurons:
\begin{equation}
dist_{IT} = \sum\limits_{i = 1}^m \sum\limits_{j = 1}^n {Feat_{ij}^2}, 
\end{equation}
where $Feat_{ij}$ is a feature number at $ij$-$th$ element of Feature Map, $m$ and $n$ are dimensions of Feature Map, $i=1,...,m$ and $j=1,...,n$. We consider that $\beta_{IT} = (dist_{IT})^{-1}$. For instance, the absolute value of parameter $\beta_{IT}$ for $IT_{RBF}(1)$ is $\approx10^{-7}$. Here notation ``(1)" indicates the number of the object stored.

The recognition procedure depends on the threshold value $\alpha$, i.e., if activation of the $IT_{RBF}$ neuron is less then $\alpha$, the actual stimulus is considered to be different from the corresponding object. We set $\alpha$ to 0.67 \cite{op1}(p. 1690) on the normalized scale. Therefore the threshold value $\alpha$ is used for detection of new objects (novelty detection) and stored objects' recognition.

In general case, a stimulus processing should be based on the \textit{predictive coding} (see next section). However, to apply predictive coding some objects should be already stored in IT.
Consider the Object 1 in Fig.~\ref{stH}. This object is regarded as the very first input to the system. Therefore, the representation of the input image is stored by the $IT_{gSOM}(1)$ neuron as a new object. This causes creation of an entire new class of $IT_{RBF}(1)$ neurons. The RBF-units of a single class densely cover the Feature Map and produce the $IT_{RBF}$ response grid.

\begin{figure}
\centering
\includegraphics[width=8.5cm]{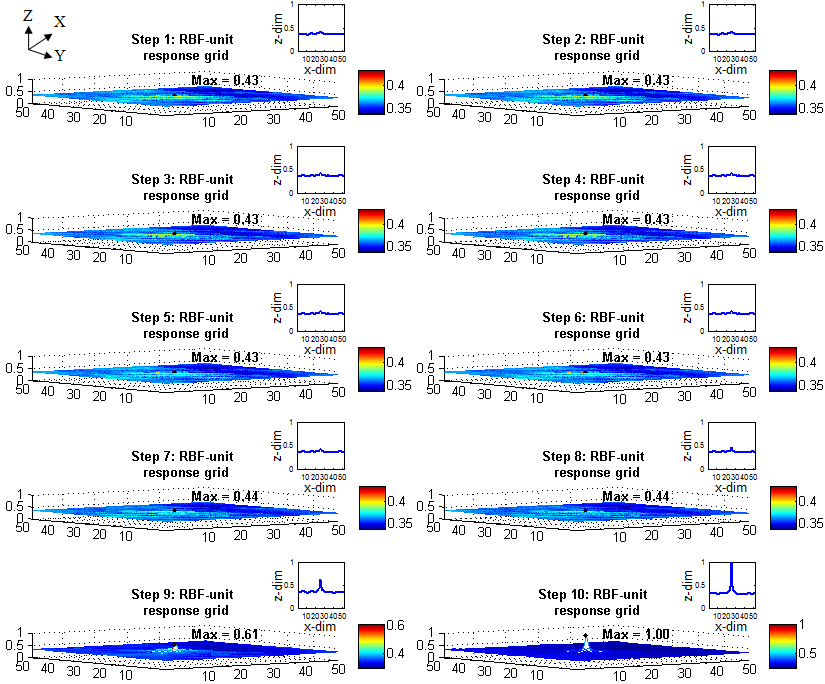}
\caption{The RBF response grids to waves of spikes. Each grid represents the responses of $IT_{RBF}$ neurons, which cover densely the Integrated Orientation Map. The center of the RBF units is Hand 1 (Object 1). The input on each step is the cumulative neuronal signals about Hand 1 delivered by waves of spikes, arrived up to this step. The maximum response is indicated by the black circle in each grid. The curve in the top-right corner of each grid indicates the section of the grid in the $XZ$-plane with fixed value of $Y$-component. This fixed value shows location of the maximum response on the $Y$-axis. $X$ and $Y$ axis correspond to horizontal ($n$) and vertical ($m$) dimensions, respectively, of the original retinal image. $Z$-axis corresponds to the magnitude of RBF responses.}
\label{h1h1} % Give a unique label
\end{figure}

\textit{Response of $IT_{RBF}$ neurons.} Now, there is a single object in IT. Here, we illustrate the response grids of $IT_{RBF}$ neurons, densely covering the Feature Map, to the sequence of waves of spike. The response grids are presented in Fig.~\ref{h1h1}.

\subsection{Predictive Coding: Interaction between V4 and IT}
Predictive coding is a widely discussed scientific paradigm \cite{kawato1,rao:bal}. It involves two stages. First, based on some preliminary data about the stimulus, the \textit{Initial Hypothesis} is generated. Then additional information about the stimulus is used to either confirm or reject the Initial Hypothesis through the process of \textit{Iterative Refinement}. Thus, formation of the Initial Hypothesis and its Iterative Refinement are two key points of the predictive coding. 

\begin{figure}
\centering
\includegraphics[width=8.5cm]{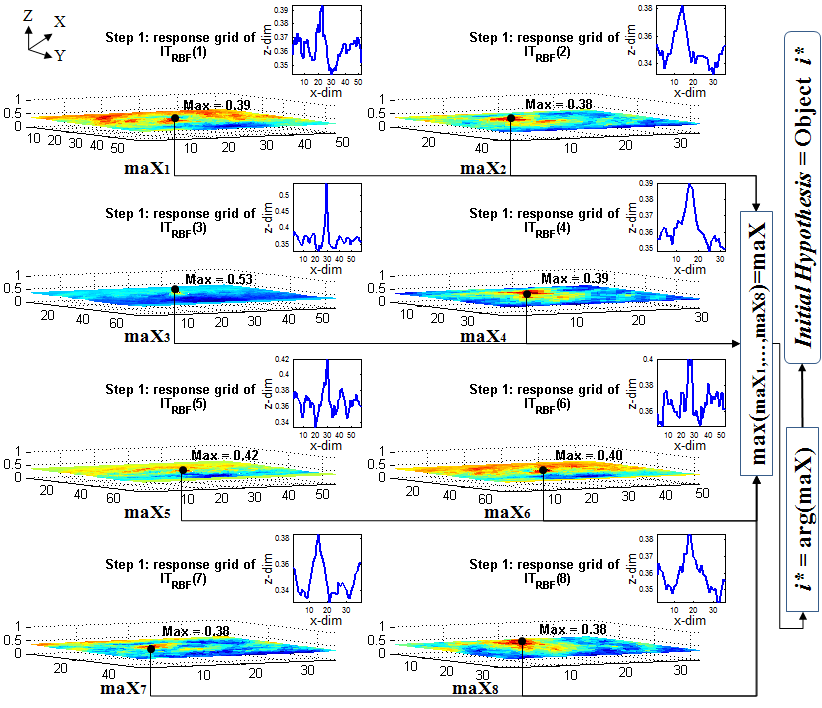}
\caption{Schema of Initial Hypothesis Generation. First the $maX_i$ responses are compared to detect, which $IT_{RBF}(i)$ neurons has the stongerst activation. Then the max value $maX$ is choosen among the $maX_i$, $i=1,...,K$. Finally, the object corresponding to the maX value is identified: $i^* = arg(maX)$. Object $i^*$ is considered as \textit{Initial Hypothesis}. Here $K=8$, $maX=0.53$ and $i^* = 3$.}
\label{initHypothesisSch} % Give a unique label
\end{figure}

\textit{Selection of the Initial Hypothesis.}
Generation of the \textit{Initial Hypothesis} is based on the information delivered by the First Wave of Spikes and the objects stored in IT. In this study, the system stores 10 objects in IT (Fig.~\ref{stH}). The steps of the \textit{Initial Hypothesis Generation} are the following: 1) the response grids for each type $i$ of $IT_{RBF}(i)$, $i=1,...,K$, $K$ is a total number of objects stored, neurons are calculated; 2) maximum activation value $maX_i$ for each grid is identified; 3) maximum value $maX$ from $maX_i$: $maX = \mathop {\max }\limits_i\{maX_i\}$; and 4) identify the number of the object, which will be considered as Initial Hypothesis: $i^* = arg(maX)$. Steps 2 and 3 are based on the $max$-$pooling$ operation performed in IT \cite{riepog}.
The object contained by $IT_{gSOM}(i^*)$, where $i^* = arg(maX)$, is considered to be the initial hypothesis for the presented stimulus. The Initial Hypothesis generation procedure is presented in Fig.~\ref{initHypothesisSch}. Here $K=8$, $maX=0.53$ and $i^* = 3$. 

\textit{Iterative Refinement.} The Initial Hypothesis is presented in a form of a Feature Map of a certain object stored in IT. When appropriate Feature Map is found, it is possible to send modulatory signal from IT back to V4. The purpose of this modulation is to amplify the activity of the neurons, which are coherent with Initial Hypothesis.

For extraction of the coherent/identical elements (neurons), the Feature Maps of the actual stimulus and the Initial Hypothesis are compared element-wise. The identical elements in both maps are localized. Thus, the neurons representing identical visual features in both maps are detected. Among the localized identical elements, the elements with activation level 1 belong to the First Wave of Spikes. Therefore they are excluded by means of inhibition. This is called to be a \textit{matching procedure}. The activation levels of remaining identical elements in Feature Map are amplified by the top-down modulatory signals from IT:
\begin{eqnarray}
\label{amp}
AL_{amp} = AL + 1; \\
\label{amp1}
AL_{amp} > 1 \rightarrow AL_{amp} = 1.
\end{eqnarray}
where $AL$ is the activation level before the modulatory top-down signal, and $AL_{amp}$ is the amplified activation level after modulatory top-down signal. 

It is possible that actual stimulus is one of the objects already stored in IT or slightly damaged with noise. Then the Initial Hypothesis has many common features with internal representation of the stimulus distributed among the waves of spikes. Therefore by means of modulation (amplification), the major amount of information will be instantly accumulated by the Second Wave of Spikes, forming ``\textit{Tsunami of Spikes}". In this case, the significant \textit{accelerating effect} will occur. We refer to such Initial Hypothesis as $coherent$. On the other hand, if the actual stimulus is not an object from IT, then the Initial Hypothesis will have few identical neurons with the waves of spikes produced by this stimulus. Consequently, only some small amount of information distributed among the waves of spikes will be shifted to the Second Wave of Spikes, and no significant accelerating effect will take place. Such Initial Hypothesis as called an $incoherent$. 

\begin{figure}
\centering
\includegraphics[width=5cm]{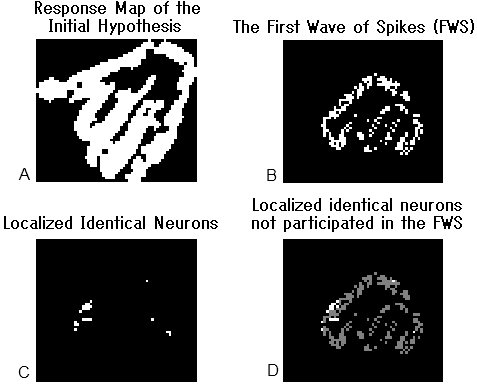}
\caption{The absence of an acceleration effect on wave propagation when coherent Initial Hypothesis does not exist. $A.$ The response map of Initial Hypothesis (Object 2). $B.$ The First Wave of Spikes of a stimulus (Object 1). $C.$ All identical elements (neurons). $D.$ The neurons not participating in the First Wave of Spikes with visual features identical to the ones of the Initial Hypothesis.}
\label{predict2} % Give a unique label
\end{figure}

\begin{figure}
\centering
\includegraphics[width=5cm]{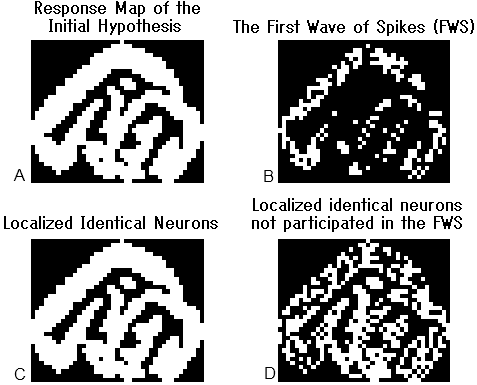}
\caption{The presence of an acceleration effect on wave propagation when a coherent Initial Hypothesis exists. $A.$ The response map of Initial Hypothesis (Object 1). $B.$ The First Wave of Spikes of a stimulus (Object 1). $C.$ All identical elements (neurons). $D.$ The neurons not participating in the First Wave of Spikes with visual features identical to the ones of the Initial Hypothesis.}
\label{predict1} % Give a unique label
\end{figure}

\begin{figure}
\centering
\includegraphics[width=8cm]{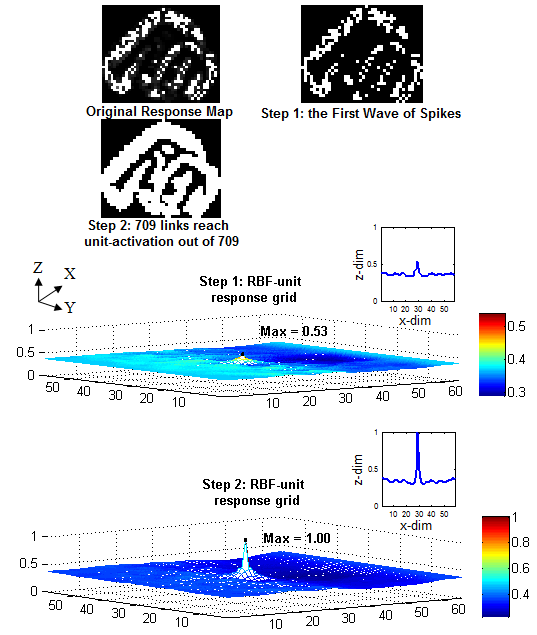}
\caption{The illutration of the Accelerating effect with waves of spikes and RBF response grids. $Top.$ The Original Response Map, the First Wave of Spikes and the Second Wave of modulated (amplified) Spikes. $Bottom.$ The response grid of $IT_{RBF}(1)$ neurons to the First Wave of Spikes and to the Second Wave of Spikes.}
\label{prcod} % Give a unique label
\end{figure}

Suggest, considering only Objects 2 and 3. We suppose that only Object 3 is stored in IT. In such a case, when Object 2 is presented as stimulus, there will be no accelerating effect (Fig.~\ref{predict2}), because the stimulus (Object 2) and RBF-center (Object 3) have only few common features. On the other hand, if Object 3 is used as a stimulus, the significant accelerating effect takes place (Fig.~\ref{predict1}), because in such a case both stimulus and RBF-center have many common features. The numerical aspects of acceleration are illustrated in Fig.~\ref{prcod}.

Summarizing, we present algorithm of Iterative Refinement process as follows: 1) after the activation levels of neurons have been modulated (amplified), the excitatory waves of spikes is sent to IT; 2) the hypothesis generation is repeated. The Initial Hypothesis is either confirmed and retained or rejected and changed; 3) the procedure of matching identical elements is repeated; 4) and modulatory top-down signals are sent back to V4. The Iterative Refinement continues until all waves of spikes reach IT. 
\section{Development: collecting new features and storing new objects}
\subsection{Towards universal dictionary of features}

Development is characterized by the dynamical change in topology of in intra- and inter-level connection. This changed involves not only extraction of new features by means of growing, but also disposal of existing rarely used features. In this section we discuss a mechanism of disposal of existing features. 

The features are extracted on the basis of the distance. On the other hand, during the processing of a particular stimulus, some features can be found more often than the others. 

To distinguish frequently used features, the counter variable $\tau$ is used. Each time feature $i$ is the best matching feature, the counter $\tau_i$ is incremented by 1. Then the relative frequency $fr_i$ of activation for each feature is calculated:
\begin{equation}
fr_{i} = \frac{\tau_i}{\sum\limits_{i = 1}^F{\tau_{i}}}, 
\end{equation}
where $i$ is a feature number, $i=1,...,F$, and $F$ a total number of features \cite{fritzke}. 

\begin{figure}
\centering
\includegraphics[width=8.5cm]{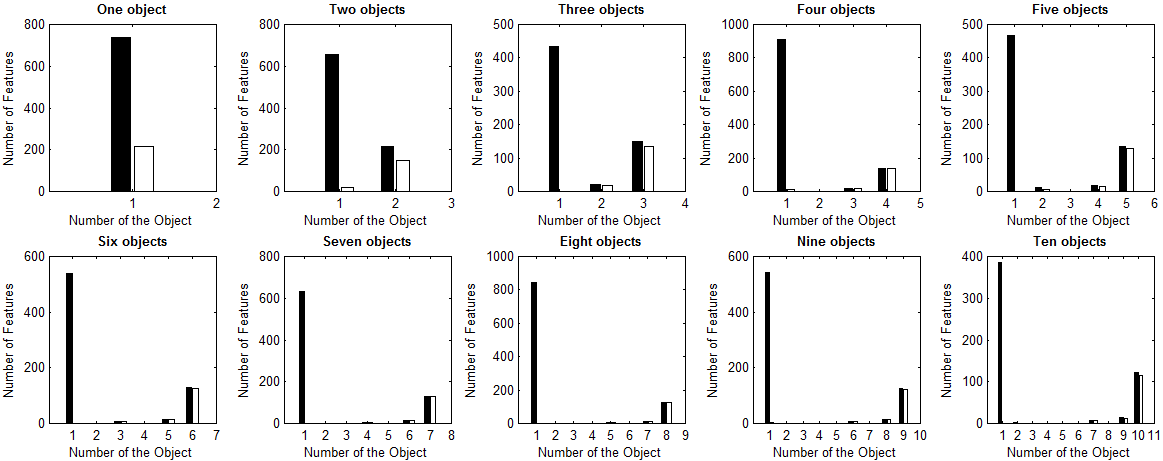}
\caption{The filled bars indicate number of all originally extracted distant enough features. The empty bars show the number of survived features with relative frequency of activation greater than chance level. The features have been extracted from ten objects. Each graph corresponds to processing of a single object. The abscissa axis indicates a number of stimuli (objects) features have survived. The ordinate axis shows the absolute number of features. For the very first stimulus, the number of features before disposal is the number of originally extracted distant enough features. For the other stimuli, the features before disposal include features, survived the previous stimuli, and the new features, extracted from the current stimulus.}
\label{evDyn} % Give a unique label
\end{figure}

Finally, we dispose the features, which have relative frequency less or equal to the chance level $1/F$. We refer to remaining features as to \textit{survived features}. The process of feature extraction and disposal is illustrated in Fig.~\ref{evDyn}. The outburst at x = 1 on abscissa axis is explained by the great number of originally extracted object specific features.

\begin{figure}
\centering
\includegraphics[width=5cm]{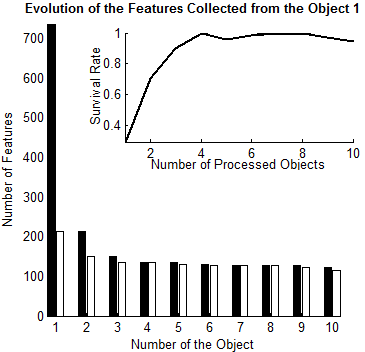}
\caption{Evolution of number of features, extracted from the very first stimulus (object). The empty bars indicate number of features survived. The filled bars show the number of features before disposal. For the very first stimulus, the number of features before disposal is the number of originally extracted distant enough features. For the other stimuli, the features before disposal are the features, which were extracted from the very first stimulus and survived all previous stimuli. The inlay illustrates dynamics of the \textit{survival rate}, calculated as ratio of number of survived features to the number of features before disposal.}
\label{FFDyn} % Give a unique label
\end{figure}

Among the features collected from ten objects, there are features that have been extracted from the very first stimulus and survived throughout the objects. We illustrate the evolution of number of such features in Fig.~\ref{FFDyn}. The inlay in Fig.~\ref{FFDyn} illustrates dynamics of \text{survival rate}. 

The features survived all ten stimuli represent the collection of features $universal$ for all the objects. Therefore, during development, extraction of universal feature intrinsic for all stimuli and disposal of stimulus specific features. This illustrates convergence of the collection of features to the \textit{Universal Dictionary of Features}\footnote{Universality is understood from the point of view of the visual experience represented by ten different objects}.
\subsection{Dynamic structure of the inter-level connections between the V4 and IT areas}
The developmental process implies two major changes: 1) the growth of number of features (in V4) and objects (in IT) together; and 2) emergence of new connections between the new neurons and the existing ones. From the very beginning , the networks in the V4 and IT levels of the system are empty. During the development, these networks are growing, thus, causing the installation of new intra- and inter-level connections. The new features are extracted and disposed in V4. In the IT module, the new object is added if the $maX$ value after predictive coding is less than threshold $\alpha$. If accelerating effect takes place, the predictive coding is terminated instantly. Therefore, emergence of both intra- and inter-level connections in our system involves both the V4 and IT areas. 
\begin{figure}
\centering
\includegraphics[width=8.5cm]{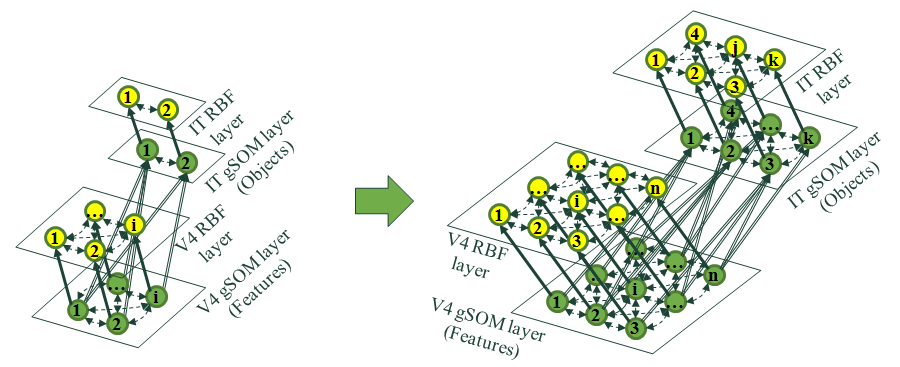}
\caption{Development of the V4 and IT levels of the system. $Left$. At some earlier stage of development, when few features ($i$ in total) and objects (2 in total) are collected, there are few neurons in gSOM and RBF layers. $Right$. On the later stage of development, number of features grows up to $n > k$, and number of objects increases up two $k>2$. This causes emergence of new numerous intra- and inter-level connections. The number of connections increases drastically.}
\label{developG} % Give a unique label
\end{figure}

The process of feature disposal takes place after the information about the Feature Map is send to the IT area by the waves of spikes. Therefore, after the extraction, collection of features in gSOM is updated, the Feature Maps of the objects stored in IT should be updated as well. This is implemented by means of substitution of previously used features for the best matching survived features (marked by double-end arrow-head dashed line in Fig.~\ref{sketch}). This substitution process takes place after the Predictive Coding. This process causes reciprocal process of updating objects' Feature Maps.

Therefore in the presented framework the developmental process is illustrated from both aspects of structure change (growing and disposal), which cause the dynamical change in topology of intra- and inter-level connections.

The schema of development of the V4 and IT areas, including intra- and inter-level connections, is presented in Fig.~\ref{developG}.
\section{Discussion and Conclusion}
We have presented a framework to model development of cortex-like visual object recognition system. This system is bases on the neurophysiological principles of information processing along the VVS. These are hierarchical organization of the visual object recognition system, propagation of waves of spikes, reciprocal interaction of different hierarchical levels, predictive coding and dynamic structure of the visual features repository. Our model also inherits major features of the RBF-units as a model of a single neuron in V4 \cite{pasucon} and IT \cite{logpog}. 

The novelty of our model is the application of growing SOMs to model the processes of feature and object extraction and storage in areas V4 and IT, respectively. We have also implemented the processes of feature extraction and feature disposal. The interplay of these processes allows to simulate dynamical changes in the structure of visual feature repository and consequent update of the objects.

We have illustrated the ability of our system to develop over time by extracting new features and storing new objects. We have shown that the predictive coding can speed up the processing time by amplifying the neuronal responses. This suggests the short processing time could be achieved in architectures with reciprocal connections. The evolution of the feature repository allows to illustrate a process of formation of \textit{Universal Dictionary of Features}. 

Therefore, our framework is a successfully working system of visual object recognition, capable of on-line development from visual experience. The development of the system takes place for each presented stimulus in two ways. First, on-line extraction of features and objects allows to model emergence of feature selectivity in V4 and object selectivity in IT. Second, the interaction between the V4 and IT growing SOM layers provides the system with ability of extracting only the features found most frequently in various stimuli, thus, collecting the \textit{universal features}.

There are many aspects for the further extension of the system: 1) the classification ability of the neurons in the Inferior Temporal cortex \cite{logpog}; 2) emergence of modular units like columnar structures \cite{korner}; 3) continuous topological structure of object representation \cite{op1} etc. One of the possible further extension beyond the visual domain is to make this framework capable of operating with multimodal signals and perform cross-modal integration.
Moreover, the modular structure of the system implies the flexibility of the system's architecture. Therefore, this framework can be used as a basis to test various biologically plausible configurations. For example, it is possible to extend the system with V2 and/or V3 areas without restrictions. 

Furthermore, the capability of predictive coding makes this framework to be applicable and extendable to model Dorsal Visual Stream for motion recognition, Superior Temporal Sulcus for biological motion recognition etc.

Moreover, this framework has a potential serve as a platform to model large-scale cortical networks like Mirror Neuron System.

\section{Acknowledgment}
The author would like to thank professor Toshio Inui, JST ERATO Asada Synergistic Intelligence Project and
Kyoto University, Graduate School of Informatics, for valuable comments and comprehensive discussions.

\def\V{\rm vol.~}
\def\N{no.~}
\def\pp{pp.~}
\def\Pot{\it Proc. }
\def\IJCNN{\it International Joint Conference on Neural Networks\rm }
\def\ACC{\it American Control Conference\rm }
\def\SMC{\it IEEE Trans. Systems\rm , \it Man\rm , and \it Cybernetics\rm }
\def\handb{ \it Handbook of Intelligent Control: Neural\rm , \it
Fuzzy\rm , \it and Adaptive Approaches \rm }

\end{document}